\documentclass{article}

\usepackage{arxiv}

\usepackage[utf8]{inputenc} 
\usepackage[T1]{fontenc}    
\usepackage{hyperref}       
\usepackage{url}            
\usepackage{booktabs}       
\usepackage{amsfonts}       
\usepackage{nicefrac}       
\usepackage{microtype}      
\usepackage{lipsum}
\usepackage{graphicx}
\usepackage{caption}
\usepackage{subcaption}
\graphicspath{ {./images/} }

\title{Orbital Collision: An Indigenously Developed Web-based Space Situational Awareness Platform\thanks{Supported by Infosys Centre For Artificial Intelligence, Indraprastha Institute of Information Technology, IIIT Delhi.}}

\author{
 Partha Chowdhury \\
 PhD Scholar \\
  Department of ECE\\
  Indraprastha Institute of Information Technology, Delhi\\
  New Delhi, 110020 \\
  \texttt{parthac@iiitd.ac.in} \\
   \And
   Harsha M \\
 PhD Scholar \\
  Department of ECE\\
  Indraprastha Institute of Information Technology, Delhi\\
  New Delhi, 110020 \\
  \texttt{harsham@iiitd.ac.in} \\
   \And
   Ayush Gupta \\
   Mtech \\
  Department of ECE\\
  Indraprastha Institute of Information Technology, Delhi\\
  New Delhi, 110020 \\
  \texttt{parthac@iiitd.ac.in} \\
  \And
   Sanat K Biswas \\
 Assistant Professor \\
  Department of ECE\\
  Indraprastha Institute of Information Technology, Delhi\\
  New Delhi, 110020 \\
  \texttt{sanat@iiitd.ac.in} \\
}

\begin{document}
\maketitle
\begin{abstract}
This work presents an indigenous web-based platform Orbital Collision (OrCo), created by the Space Systems Laboratory at IIIT Delhi, to enhance Space Situational Awareness (SSA) by predicting collision probabilities of space objects using Two-Line Element (TLE) data. The work highlights the growing challenges of congestion in the Earth’s orbital environment, mainly due to space debris and defunct satellites, which increase collision risks. It employs several methods for propagating orbital uncertainty and calculating the collision probability. The performance of the platform is evaluated through accuracy assessments and efficiency metrics, in order to improve the tracking of space objects and ensure the safety of the satellite in congested space.

\end{abstract}


\section{Introduction}
With the rapid expansion of space-based infrastructure supporting global communication, navigation, and scientific exploration, the safe and sustainable operation of satellites has become a critical concern. However, the Earth's orbital environment is becoming increasingly congested due to the accumulation of space debris and inactive satellites, significantly elevating the risk of in-orbit collisions. These threats underscore the urgent need for advanced Space Situational Awareness (SSA) systems capable of accurately monitoring and predicting the dynamics of space objects.

This paper presents Orbital Collision (OrCo), a Python-based web platform developed by the Space Systems Laboratory at IIIT Delhi to enhance SSA capabilities. OrCo integrates real-time orbital data retrieval, uncertainty propagation using Monte Carlo methods, and collision probability estimation within an interactive and accessible web interface. The platform supports intuitive visualization tools for assessing conjunction risks and analyzing space object behavior.

A comprehensive evaluation of OrCo’s architecture, algorithms, and computational performance is conducted, including benchmarking against data from the Space Surveillance Network. The results demonstrate OrCo’s accuracy, scalability, and operational viability, positioning it as a valuable contribution to autonomous space traffic management. This work supports creating open and efficient systems to track space objects, and ensure sustainable space activities over the time.

The rest of the paper is organized as follows: Section 2 discusses the background and the purposes of OrCo. Section 3 describes the workflow and methodologies used in this study. Section 4 discuss the results. Section 5 concludes the paper and outlines future directions.

\section{What is Orbial Collision (OrCo)?}

OrCo (Orbital Collision) is a web-based platform indegenously
developed by the Space Systems Laboratory at IIIT Delhi, with support from the Infosys Centre for Artificial Intelligence, aimed at advancing real-time Space Situational Awareness (SSA). The platform retrieves orbital data in the form of Two-Line Elements (TLEs) sourced from \href{www.space-track.org}{space-track.org} and employs the SGP4 propagator for efficient position forecasting. To address uncertainties in orbital dynamics, OrCo utilizes a stochastic framework based on Monte Carlo sampling to generate probabilistic future states and estimate collision probabilities through analysis of overlapping uncertainty regions. In addition to its computational capabilities, OrCo features interactive visualization tools that display conjunction probabilities and covariance ellipsoids, while enabling filtering based on orbital parameters. 

\begin{figure}[ht]
    \centering
    \includegraphics[width=0.4\linewidth]{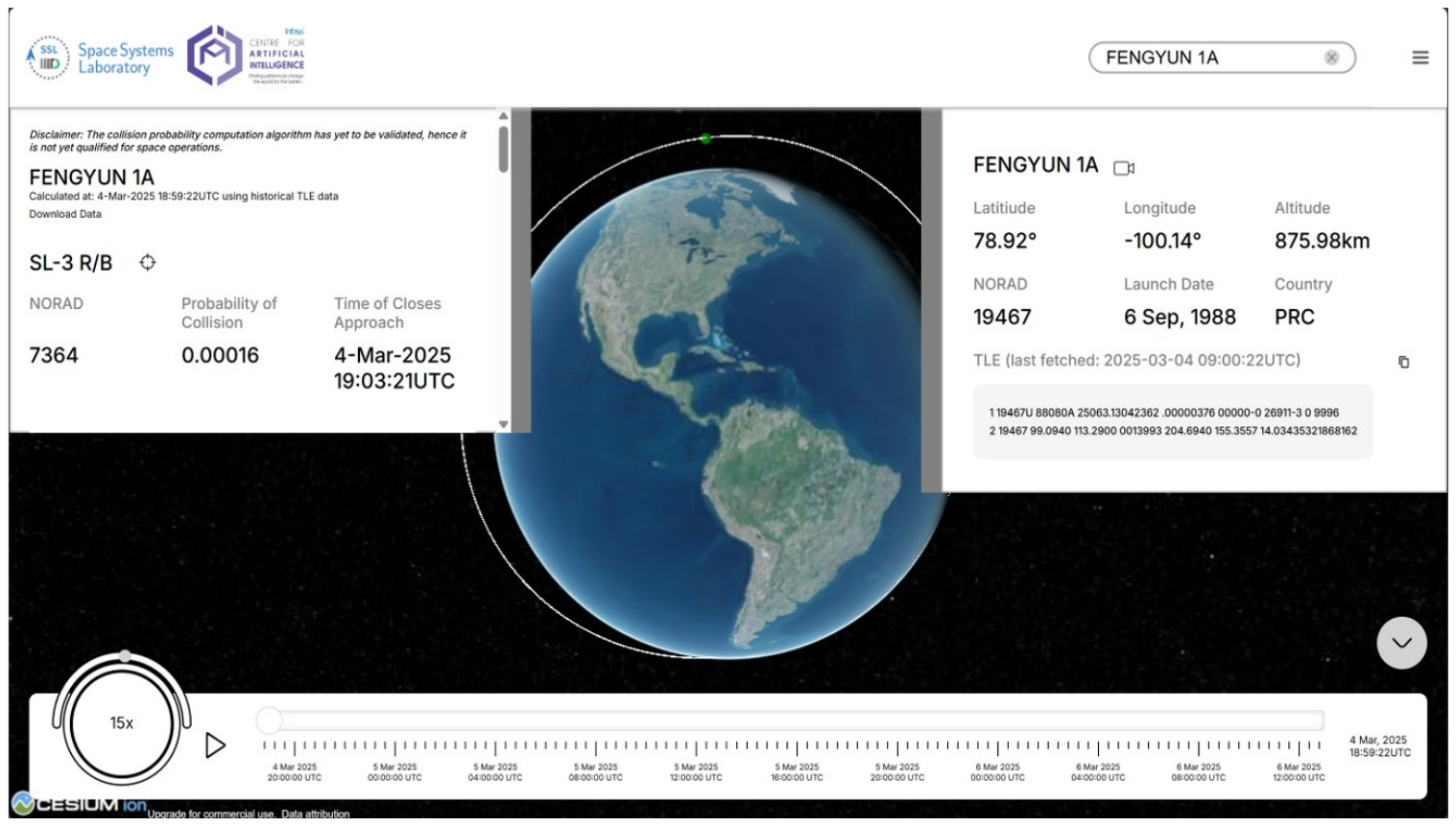}
    \caption{OrCo Website}
    \label{fig:enter-label}
\end{figure}

\section{Methodology}

\begin{figure}[ht]
    \centering
    \includegraphics[width=0.5\linewidth]{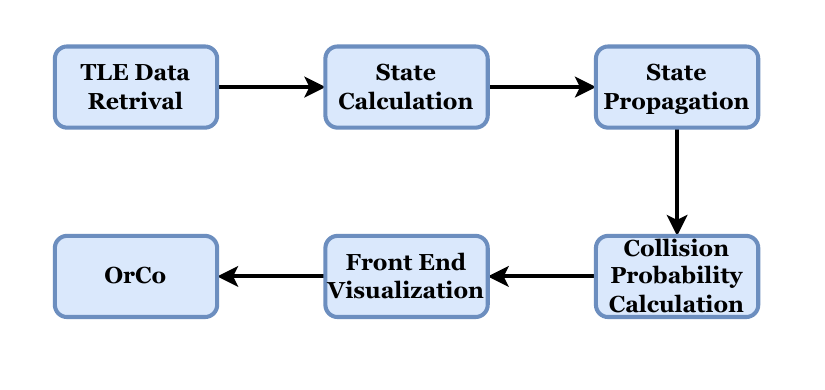}
    \caption{Workflow of OrCo}
    \label{fig:Work-Flow}
\end{figure}

The OrCo follows a streamlined and modular workflow (fig: \ref{fig:Work-Flow}) designed for real-time collision risk assessment. It begins with the retrieval of TLE data from Space-Track.org, which provides the necessary orbital parameters for each space object. These parameters are used in the state calculation phase to compute the initial positions and velocities of objects using the SGP4 propagator. Next, the platform performs state propagation, employing methods such as Monte Carlo, ESPT, and Advanced ESPT to simulate the evolution of object states while accounting for positional uncertainty. Based on the propagated states, collision probability is calculated using algorithms like Monte Carlo, Patera’s, and Alfano’s methods, which estimate the likelihood of a conjunction at the time of closest approach. Finally, the results are processed through a front-end visualization interface, where users can interactively analyze conjunction risks via scatter plots, covariance ellipsoids, and object filtering tools. This integrated workflow enables accurate, efficient, and user-accessible space traffic monitoring.

\subsection{State Propagation}
The state propagation in OrCo is designed to account for uncertainties in the future positions of space objects using three complementary approaches: Monte Carlo Propagation, Extrapolated State Propagation Technique (ESPT), and Advanced ESPT (AESPT). These methods start with the ingestion of orbital data in the form of TLEs and progress through different pathways to generate probabilistic future states.

Monte Carlo Propagation involves sampling the initial state (position and velocity) from a predefined distribution, usually Gaussian. Each sample is then propagated forward in time using the SGP4 model. This generates a cloud of possible future positions, enabling statistical analysis of conjunction risk. This method is simple and widely used for its clarity but can be computationally intensive for large sample sizes. But Extrapolated State Propagation Technique (ESPT)
aims to reduce the computational burden of Monte Carlo methods while preserving statistical fidelity. In ESPT, instead of propagating thousands of individual samples, a representative set of extrapolated states is computed using statistical moment approximations. These states are analytically extrapolated over time using simplified uncertainty evolution models. ESPT significantly speeds up the process by avoiding full-scale propagation for each sample while maintaining a reliable estimation of state dispersion. Moreover, the Advanced ESPT enhances the standard ESPT by incorporating adaptive correction mechanisms and higher-order extrapolation models. It adjusts the propagation process based on system dynamics and time-dependent uncertainty growth. AESPT is particularly effective for longer prediction windows or when precise estimation of uncertainty bounds is necessary. It offers a balance between computational efficiency and dynamic accuracy, outperforming ESPT in scenarios with complex orbital interactions.

\begin{figure}[ht]
    \centering
    \begin{subfigure}{.45\textwidth}
    \centering
    \includegraphics[width=\linewidth]{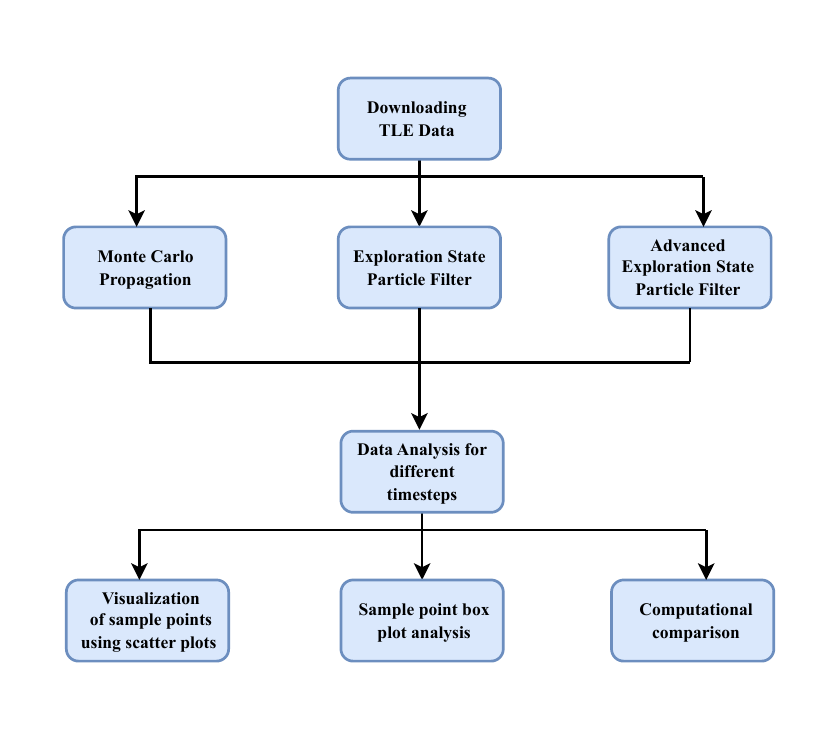}
    \caption{State Propagation}
    \end{subfigure}
    \begin{subfigure}{.45\textwidth}
    \centering
    \includegraphics[width=\linewidth]{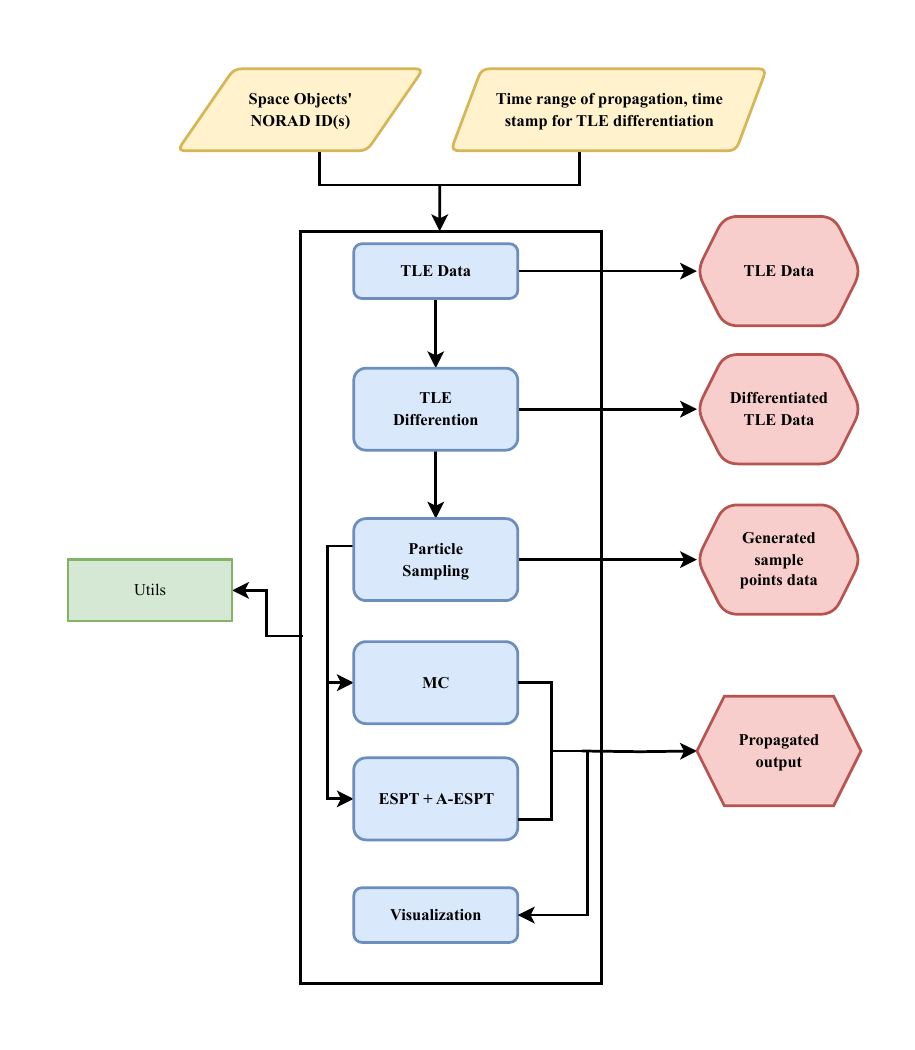}
    \caption{Simulation Module Architecture}
    \end{subfigure}
\end{figure}

After the propagation through all methods for different timestamps stage the outputs were evaluated and visualised and further used for calculation of collision probability.
\subsection{Calculation of Collision Probability}
Accurate collision probability estimation between Resident Space Objects (RSOs) is a critical aspect of Space Situational Awareness (SSA), enabling effective prediction and mitigation of conjunction events. This process involves representing positional uncertainties through three-dimensional probability density functions, typically Gaussian, and integrating over a control volume defined by the combined hard-body sphere of the objects. Collision probability is generally classified as either instantaneous or cumulative. The instantaneous probability quantifies the risk at a specific moment, most commonly at the time of closest approach (TCA), while the cumulative probability evaluates the likelihood over a specified time interval. For short-term, high-velocity encounters, the computation is often simplified by projecting the problem onto a two-dimensional encounter plane normal to the relative velocity vector, thus reducing complexity without compromising accuracy.

Three methods has been used in OrCo for calculating collision probability: 1) Monte Carlo Method, 2) Patera's Method, 3) Alfano's Method. Monte Carlo Simulation generates a number of random samples of object positions, propagates them to TCA, and counts the number of collisions within the threshold radius. While accurate, this method is computationally intensive. On the other hand, Patera’s Method transforms the 3D Gaussian PDF into a 2D form and integrates around the boundary of the hard-body area using a single integral with an exponential function. It supports irregular object shapes and offers computational efficiency. Alfano’s Method simplifies the double integral into a single one using error functions and Simpson’s numerical rule. It assumes spherical objects and linear motion, making it suitable for fast and reliable estimation under typical short-term encounters. Together, these methods allow OrCo to balance between computational efficiency and predictive accuracy, enabling practical and scalable conjunction analysis for real-time space operations.

\section{Results and Discussions}
The OrCo platform was rigorously evaluated across key functional dimensions, including state propagation accuracy, collision probability estimation, simulation fidelity, and computational performance. Three propagation techniques—Monte Carlo, ESPT, and Advanced ESPT—were tested, with ESPT-based methods offering significant reductions in runtime while preserving sufficient accuracy for short-term predictions.
For collision probability assessment, Monte Carlo served as the benchmark due to its statistical robustness. Patera’s method provided comparable accuracy with improved efficiency through contour integration, while Alfano’s method demonstrated the highest computational speed. Both "Sure Shot" and "Miss Case" simulations validated the expected behavior of decreasing collision probability with increased positional uncertainty or miss distance. Moreover to check computational efficiency we have analysed the performance of OrCo with spacetrack.org for different space objects with respect to the probability of collision and time of closest approach in tables \ref{table:1} and \ref{table:2}.

\begin{table}
\centering
\begin{tabular}{ |c|c|c| } 
 \hline
 Platform\  & TCPA\  & Collision Probability\  \\
 \hline\hline
 www.orco.edu.in & 2025-01-20  11:58:43 & 0.006 \\ 
 \hline
 www.space-track.org & 2025-01-20 15:28:45 & 0.0002382545 \\ 
 \hline
\end{tabular}
\medskip
\caption{ Collision analysis between 35364 (METEOR 2-17 DEB), 60886 (CZ-6A DEB)}
\label{table:1}
\end{table}
\begin{table}
\centering
\begin{tabular}{ |c|c|c| } 
 \hline
 Platform\  & TCPA\  & Collision Probability\  \\
 \hline\hline
 www.orco.edu.in & 2025-02-22 5:44:25 & 0.001 \\ 
 \hline
 www.space-track.org & 2025-02-22 5:48:02 & 0.0005973233 \\ 
 \hline
\end{tabular}
\medskip
\caption{Collision analysis between 60698(CZ-6A DEB), 30405 (FENGYUN 1C DEB)}
\label{table:2}
\end{table}

Hence we can validate the performance of our platform through some comparison which produces almost nearby results within a certain threshold of spacetrack.org
\section{Conclusion}

This study presented the design, implementation, and evaluation of OrCo, a web-based platform for real-time Space Situational Awareness. By integrating automated TLE data ingestion, efficient state propagation methods, and multiple collision probability algorithms, OrCo provides accurate and scalable collision risk assessments. Through extensive simulations and algorithm comparisons, the platform demonstrated strong performance in both predictive accuracy and computational efficiency. Its intuitive visualization tools further enhance user interaction and decision-making. Overall, OrCo offers a robust and accessible solution for monitoring orbital safety in an increasingly congested space environment.

%
%
%
%

\end{document}